\documentclass[conference]{IEEEtran}

\usepackage[pdftex]{graphicx}

\usepackage[cmex10]{amsmath}

\usepackage[caption=false,font=footnotesize]{subfig}

\usepackage{url}
\usepackage{xcolor}
\usepackage{supertabular}
\usepackage{arydshln}

\begin{document}
\title{Exploring the influence of scale on artist attribution}

\author{\IEEEauthorblockN{Nanne van Noord and Eric Postma}
\IEEEauthorblockA{Tilburg center for Communication and Cognition, School of Humanities \\
Tilburg University, The Netherlands\\
Email: \{n.j.e.vannoord, e.o.postma\}@tilburguniversity.edu}}


\maketitle

\begin{abstract}
    Previous work has shown that the artist of an artwork can be identified by use of computational methods that
    analyse digital images. However, the digitised artworks are often investigated at a coarse scale discarding
    many of the important details that may define an artist's style. In recent years high resolution images of artworks
    have become available, which, combined with increased processing power and new computational techniques, allow us
    to analyse digital images of artworks at a very fine scale. In this work we train and evaluate a Convolutional
    Neural Network (CNN) on the task of artist attribution using artwork images of varying resolutions. 
    To this end, we combine two existing methods to enable the application of high resolution images to CNNs. 
    By comparing the attribution performances obtained at different scales, we find that in most cases finer scales are beneficial to the attribution performance, 
    whereas for a minority of the artists, coarser scales appear to be preferable. 
    We conclude that artist attribution would benefit from a multi-scale CNN approach which vastly expands the possibilities for computational art forensics.
\end{abstract}

\IEEEpeerreviewmaketitle

\section{Introduction}
Convolutional Neural Networks (CNNs) are the current state-of-the-art classifiers for many image processing tasks \cite{Krizhevsky2012,Bengio2012a,LeCun2015}.
CNNs provide efficient tools for the automatic analysis of artworks by automatically creating visual filters \cite{Bengio1995} that respond to artist-specific characteristics \cite{Noord2015}. A trained CNN consists of multiple layers of filters, ranging from elementary general filters in the input layers to complex task-dependent filters in the output layers. A CNN trained on the task of artist attribution, performs a hierarchical decomposition of an artwork allowing it to  recognise very complex visual characteristics. In our recent work we used a CNN to achieve state-of-the-art results on a large-scale artist attribution task \cite{Noord2015}. An important limitation of our study was that the analysis was performed at a coarse scale, i.e., with a minimum resolution of $1.0-0.5$ pixels per mm.  The spatial level of analysis corresponds roughly to that of a person with normal vision ($20\//20$) examining the artwork at a distance of about $3.5-7.0$ meters! Other work on artist attribution relied on more detailed scales of analysis: $2$ pixels per mm \cite{Hughes2010a}, $7.7$ pixels per mm \cite{Johnson2008, VanderMaaten2010}, and  even $10$ pixels per mm \cite{Taylor2007}. The latter scale is comparable to viewing an artwork at a distance of $34$ cm, which seems more appropriate for the detailed analysis of artworks required for artist attribution \cite{Hendriks2008}.

The impact of varying the spatial level of analysis has never been examined in the context of computational artist attribution. The goal of this paper is to determine if and how the spatial scale of analysis (image resolution) affects artist attribution with using a CNN.
Although on the one hand, doubling the ratio of pixels per spatial unit is expected to beneficial because it reveals more visual characteristics, on the other hand, the increase of visual detail may obscure coarser visual information.  In order to obtain a fair comparison of attribution performances at different scales of analysis, we propose a CNN variant that is capable of dealing with images of arbitrary sizes. 

The remainder of this paper is organised as follows. In Section~\ref{sec:method} we describe a solution to two problems that hamper the application of CNNs to high-resolution images. In Section~\ref{sec:results} the results of the experiments are presented. We discuss the implications of using this approach and the influence of image resolution on attribution performance in Section~\ref{sec:discussion}. Finally, Section~\ref{sec:conclusion} concludes by stating that using a CNN for the analysis of multi-scale analysis of artworks, represents a fruitful avenue for artist attribution.

\section{Discovering detailed characteristics}
\label{sec:method}
Applying CNNs to multi-resolution images is hampered by two problems. The first problem is that CNNs require input images of fixed size, whereas artworks vary in their sizes. The second problem is the insurmountable computational demand imposed by very high-resolution images as CNN inputs. In what follows, we discuss each problem and its solution separately.

\subsection{Fixed input size problem} 
Many implementations of CNNs use non-convolutional (or fully-connected) output layers, mapping the result of the convolutions to a single value per class. Fully-connected output layers have been used with great success in, for example, the AlexNet architecture \cite{Krizhevsky2012}  which was used in the ImageNet challenge \cite{Russakovsky}. A restriction of the fully-connected output layers is that they constrain the CNN to have an input of fixed size, because of the fixed dimensions of the fully-connected layers. Fully Convolutional Networks (FCNs) \cite{Sermanet2013, Long2014} lift this restriction by using convolutional output layers, rather than fully connected ones. FCNs allow inputs of an arbitrary size. In our study we use an FCN to deal with the first problem.

\subsection{Computational demand problem} 
The sheer number of pixels of very high-resolution images make it computationally infeasible to train a FCN directly \cite{Hou2015}. FCNs are capable of efficiently giving predictions for large images by using a final convolutional layer which has a filter for each class followed by a pooling layer that summarises the predictions of the convolutional layer over the entire input image \cite{Sermanet2013, Long2014}. Using this approach it is possible to efficiently obtain accurate predictions for large images, but because the final convolutional layer produces predictions for many overlapping image regions the computational cost for training a FCN still increases quadratically with the input size. The computational cost can be reduced with patch-wise training. Rather than training the FCN on the entire image, it is trained on multiple smaller patches. The benefit of patch-wise training is that it allows for fine-grained image analysis without incurring insurmountable computational costs. This allows us to to train and evaluate FCNs on images with a pixel to mm ratio of $5$ and upwards.

\subsection{Solution to the two problems}
Our combination of FCN and patch-wise training provides a solution to the two problems. We achieve predictions at the level of individual pixels during testing. Furthermore, we show that this approach can be used to achieve state-of-the-art performance on an artist attribution task of artwork images with an increased resolution.

The experiments in this paper are focused on uncovering to what extent the addition of spatial detail affects artist attribution performance. For this purpose an initial experiment is performed with the $256 \times 256$ images dataset, as these are comparable in size to those used in \cite{Mensink2014, Noord2015}, which will form the baseline against which the performance of the subsequent experiments will be compared. 

The subsequent experiments will be performed on the $512 \times 512$, $1024 \times 1024$, and $2048 \times 2048$ image datasets. Where each increase in size should give a better impression of how the network deals with the increase in available details and data. 

\section{Experimental setup}
\label{sec:exp}
This section describes the setup of the artist attribution experiments with images varying from medium to high-resolution. The setup consists of a specification of the FCN architecture, the dataset, the evaluation, and the training parameters.

\subsection{Architecture}
The FCN architecture used in this paper is based on the architectures described in \cite{Krizhevsky2012,Lin2013} following modifications described in \cite{springenberg2015} as to replace the pooling layers with convolutional layers, making the network fully convolutional. A detailed description of the network can be found in Table~\ref{tab:netarch}, where conv-$n$ denotes a convolutional layer with $f$ filters with a size ranging from $11 \times 11$ to $1 \times 1$. The stride indicates step size of the convolution in pixels, and the padding indicates how much zero padding is performed before the convolution is applied. Conv-p$n$ are identical to conv-$n$ layers, except that the filters are replaced by single weights, this layer type is introduced based on the recommendations in \cite{Lin2013}, and performs an abstraction of each local patch before the patches are combined into a higher level representation. Conv-pool$n$ layers replace traditional pooling layers that might perform max or average pooling by performing convolution with a stride of $2$ in both directions, returning an output map that is a factor $4$ smaller than the input. By replacing all pooling layers with convolutional layers the network becomes suitable for Guided Backpropagation \cite{springenberg2015}, which is used to 
create visualisations of the parts of the input images that are most discriminative for a certain artist.

\begin{table}[!t]
\renewcommand{\arraystretch}{1.3}
\caption{Network architecture as used in this paper, conv$n$ are convolutional layers, whereas conv-p$n$ are based on \cite{Lin2013} and conv-pool$n$ on \cite{springenberg2015}. During training a $224 \times 224$ pixels crop is used, the testing is performed on the entire input image (of $256 \times 256$ up to $2048 \times 2048$).}
\label{tab:netarch}
\centering
\begin{tabular}{c|c|c|c}
Layer & Filters & Size, stride, pad & Description \\
\hline
Training Data & - & $224 \times 224$, -, - & RGB image crop \\
\hdashline
Testing Data & - & Entire image, -, - & Full RGB image \\
\hline
conv1 & $96$ & $11 \times 11$, 4, 0 & ReLU \\ 
conv-p1 & $96$ & $1 \times 1$, 1, 0 & ReLU \\
conv-pool1 & $96$ & $3 \times 3$, 2, 1 & ReLU \\ 
\hline 
conv2 & $256$ & $5 \times 5$, 1, 2 & ReLU \\
conv-p2 & $256$ & $1 \times 1$, 1, 0 & ReLU \\ 
conv-pool2 & $256$ & $3 \times 3$, 2, 0 & ReLU \\
\hline
conv3 & $384$ & $3 \times 3$, 1, 1 & ReLU \\
conv-p3 & $384$ & $1 \times 1$, 1, 0 & ReLU \\
conv-pool3 & $384$ & $3 \times 3$, 2, 0 & ReLU + Dropout ($50\%$) \\
\hline
conv4 & $1024$ & $1 \times 1$, 1, 0 & ReLU \\
conv5 & $1024$ & $1 \times 1$, 1, 0 & ReLU \\
conv6 & $210$ & $1 \times 1$, 1, 0 & ReLU \\
\hline
global-pool & - & - & Global average \\
softmax & - & - & Softmax layer \\
\end{tabular}
\end{table}

\subsection{Dataset}
\begin{figure}[!t]
\centering
\includegraphics[width=2.5in]{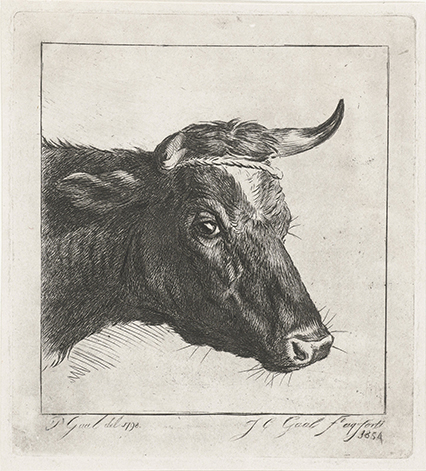}
\caption{Digital photographic reproduction of \textit{`Kop van een koe met touw om de horens'} by Jacobus Cornelis Gaal.}
\label{fig:mopshond}
\end{figure}

The dataset consists of $58,630$ digital photographic reproductions of print artworks by $210$ different artists retrieved from the collection of the Rijksmuseum, the Netherlands State Museum. These artworks were chosen by selecting all prints made on paper by a single artist, without collaboration, who were indicated to be the creator of the artwork. Further selection criteria were that the image was in the public domain and that the artist had made at least $96$ artworks that adhere to the previously mentioned criteria. This ensured that there were sufficient images available from each artist to learn to recognise their work. An example of a print from the Rijksmuseum collection is shown in Figure~\ref{fig:mopshond}.

\begin{figure}[!t]
\centering
\includegraphics[width=3in]{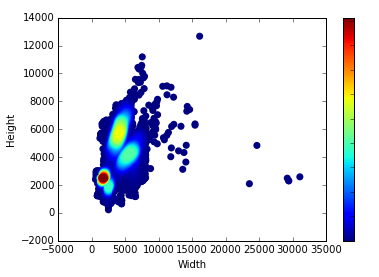}
\caption{Scatter plot of image sizes, where each point represents an artwork and the colour indicates the density in the area around that point. The images cluster around two peaks, the first peak of almost $10,000$ images  clusters around $2,500$ pixels, and the second peak between $5,000$ and $6,000$ pixels. Additionally, there are $123$ images with a dimension (width or height) larger than $10,000$ pixels.}
\label{fig:sizedist}
\end{figure}

There is a large degree of variation in the sizes and resolutions of the images in the Rijksmuseum collection, due to the physical dimensions of artworks and the resolutions of their digital reproductions. Figure~\ref{fig:sizedist} shows a scatter plot of the horizontal image sizes (widths) and vertical image sizes (heights) to illustrate the variation in size and resolution of the artworks. While previous approaches have dealt with such variations by resizing all images to a single size, it confounds image resolution with physical resolution. 
Normalising the images to obtain fixed pixel to mm ratios would result in a loss of visual detail. Hence, we take the variation in scales and resolutions for granted.
To compare the impact of image resolution on artist attribution, four version of the dataset are created from high-resolution versions of the images:  $256 \times 256$, $512 \times 512$, $1024 \times 1024$, and $2048 \times 2048$ pixels. 

The dataset is divided into a training ($70\%$), validation ($10\%$), and test set ($20\%$). The training set is used to
train the network, the validation set is used to optimise the hyperparameters, and the evaluation set is used to estimate the prediction performance. 
All results reported in this paper are based on the test set.



\subsection{Training parameters}
All networks were trained using an effective training procedure (cf. \cite{Krizhevsky2012}), with the values of the learning rate, momentum, and weight decay hyperparameters being $10^{-2}$, $0.9$, and $5 \cdot 10^{-4}$ respectively. Whenever the error on the validation set stopped decreasing the learning rate was decreased by a factor $10$. To deal with the increased size of the images the number of images per batch was adjust to fit into memory.

All training was performed on a NVIDIA Tesla K20 graphics card using the Caffe framework \cite{jia2013caffe}.

\subsection{Evaluation}
The evaluation is performed on the whole images as the FCN makes it unnecessary to take crops. The prediction for an image
is the average over many predictions for all input regions, resulting in a single prediction for the entire image. The performance on all experiments is reported using the Mean Class Accuracy (MCA), which is the average of the accuracy scores obtained per artist. We report the MCA because it is not sensitive to unbalanced classes and it allows for a comparison of the results with those reported in \cite{Mensink2014, Noord2015}.

Additionally we report the pair-wise correlations between the Class Accuracy (CA) for each artist for the four datasets at different scales. The correlation between the 
results on two different scales indicates how similar the performance is for individual artists for two image resolutions. A high correlation between two scales 
indicates that the attributions of an artist are largely the same at both scales, whereas a low correlation indicates that the artworks of an artists are classified differently at the two scales. 

\section{Results}
\label{sec:results}
Table~\ref{tab:results} lists the prediction results of the experiments on the four image datasets. The results obtained for high-resolution images ($1024 \times 1024$ and $2048 \times 2048$) outperform those reported on low-resolution images of at most $500 \times 500$ and $256 \times 256$ in \cite{Mensink2014} and \cite{Noord2015} respectively. It is important to note that while the same data source was used, the way in which the data was selected could introduce small performance differences. Nonetheless, as is evident from the results, increasing the resolution from $256 \times 256$ two-, three-, or fourfold is beneficial to the artist-attribution performance.

\begin{table}[!t]
\renewcommand{\arraystretch}{1.3}
\caption{Mean Class Accuracies of artist attribution experiments for four image resolutions.}
\label{tab:results}
\centering
\begin{tabular}{cc}
Image size & MCA \\
\hline
$256 \times 256$ & $67.0$ \\
$512 \times 512$ & $73.4$ \\
$1024 \times 1024$ & $73.1$ \\
$2048 \times 2048$ & $72.9$ \\
\hline
\end{tabular}
\end{table}

The MCAs obtained for the $512 \times 512$, $1024 \times 1024$, and the $2048 \times 2048$ sets are very similar. This suggests that there is a ceiling in performance and that further increasing the resolution would not help to improve the performance. By comparing the performance per artist it becomes apparent that the underlying pattern is more subtle. Whereas most artists seem to benefit from increasing the resolution, many seem to suffer from it. In Table~\ref{tab:corresults} the correlations between the pairs of results are shown. These correlations show that the performances obtained on the $512$ versus $256$ sets are most similar, and the $256$ versus $2048$ are most dissimilar, but that there is no single scale that is optimal for all artists.

\begin{table}[!t]
\renewcommand{\arraystretch}{1.3}
\caption{Correlations between results per artist for each image resolution}
\label{tab:corresults}
\centering
\begin{tabular}{lrrrr}
\hline
{} &  256 &  512 & 1024 &  2048 \\
\hline
256 &    1.00 &    0.71 &   0.44 &   0.29 \\
512 &    0.71 &    1.00 &   0.51 &   0.29 \\
1024  &    0.44 &    0.51 &   1.00 &   0.67 \\
2048  &    0.29 &    0.29 &   0.67 &   1.000 \\
\hline
\end{tabular}
\end{table}

Comparing the performances for each artist as a function of image resolution, three trends become apparent: increasing, decreasing, and invariant. The counts of how the performance per artists artists fits one of these trends are shown in Table~\ref{tab:summary}. For most artists the performance increases when the resolution is increased, but at the same time there are $82$ artists for whom it is actually better to perform the analysis at a lower scale.

\begin{table}[!t]
\renewcommand{\arraystretch}{1.3}
\caption{Analysis of the trends for the results per artists for different scales.}
\label{tab:summary}
\centering
\begin{tabular}{lcp{.4\linewidth}}
\hline
Trend &  Number of Artists & Description \\
\hline
Increasing &    119 & Increasing performance on higher resolution images. \\
Decreasing &    82  & Decreasing performance on higher resolution images. \\
Invariant &    9 & No discernible difference between scales. \\
\hline
\end{tabular}
\end{table}

\begin{figure}[!t]
  \centering
  \subfloat[Art work at $256 \times 256$]{
    \includegraphics[width=.45\linewidth]{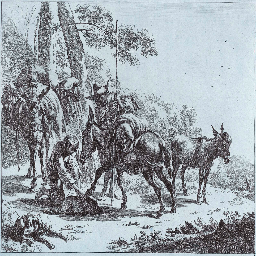}
  }
  \subfloat[Activation]{
    \includegraphics[width=.45\linewidth]{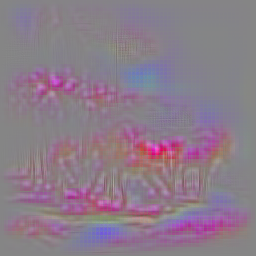}
  }
  \hfil
  \subfloat[Art work at $512 \times 512$]{
    \includegraphics[width=.45\linewidth]{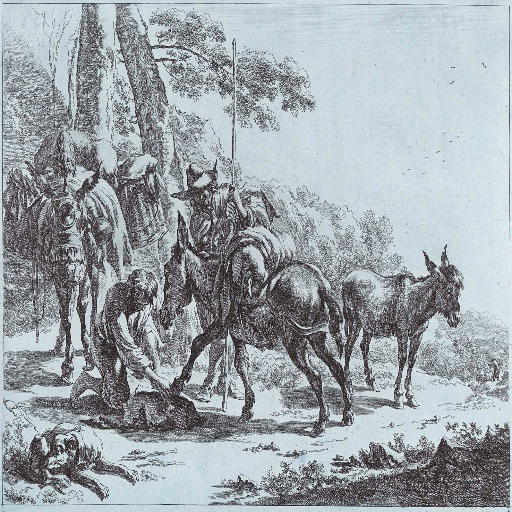}
  }
  \subfloat[Activation]{
    \includegraphics[width=.45\linewidth]{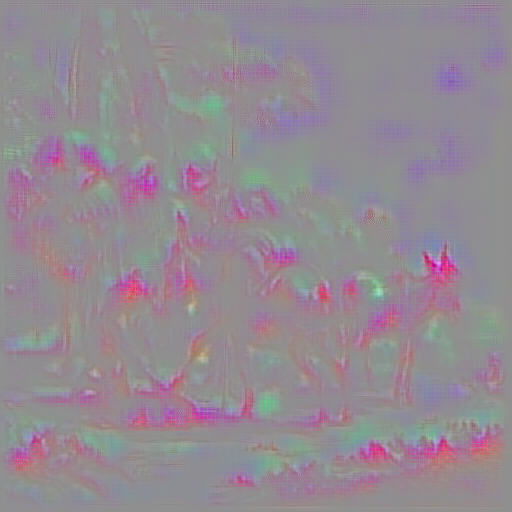}
  }
  \hfil
  \subfloat[Art work at $1024 \times 1024$]{
    \includegraphics[width=.45\linewidth]{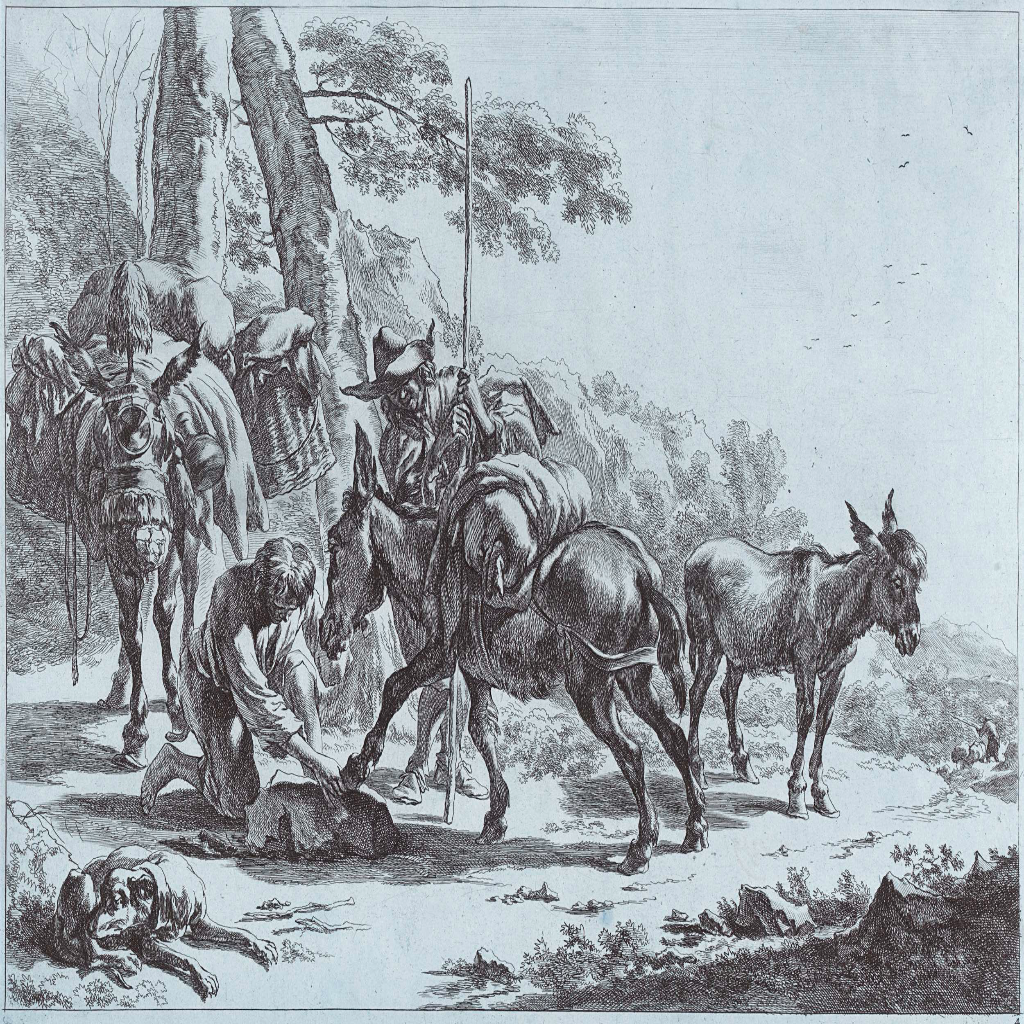}
  }
  \subfloat[Activation]{
    \includegraphics[width=.45\linewidth]{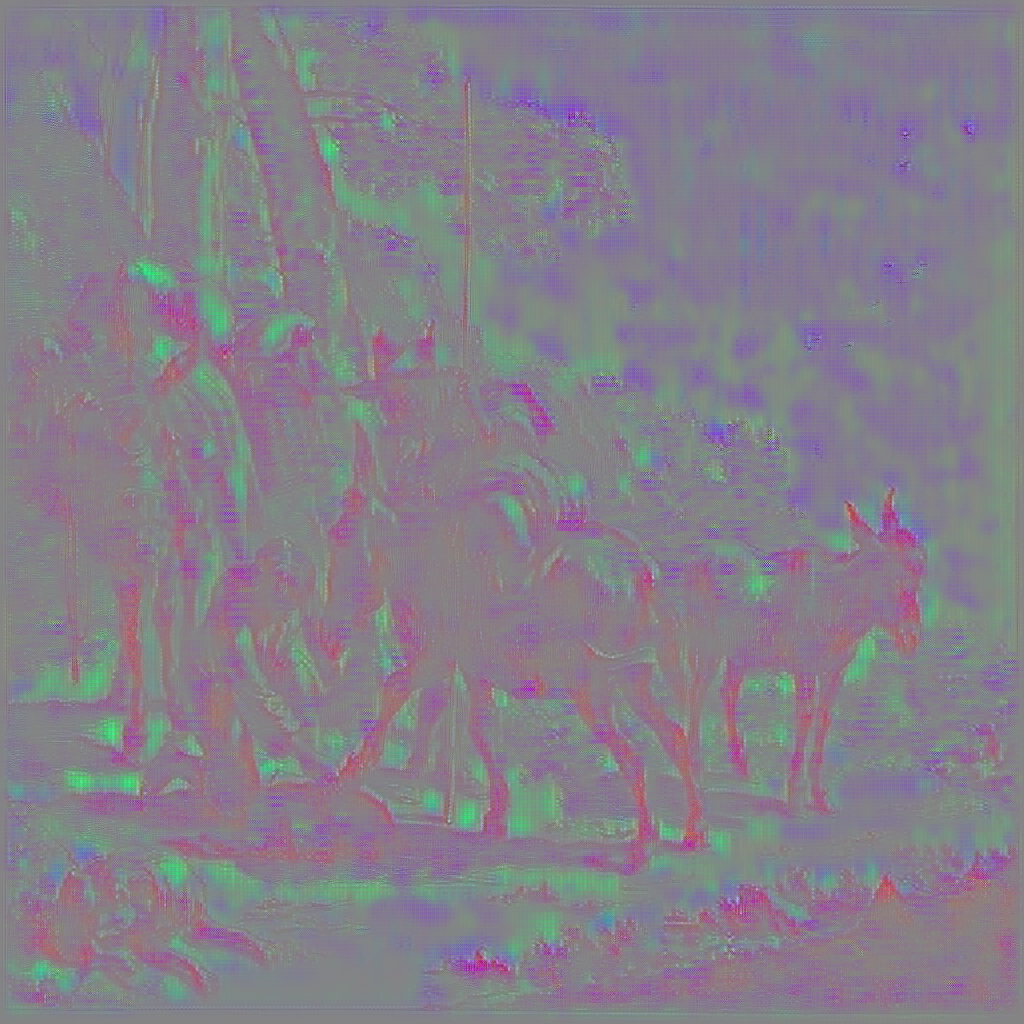}
  }
  \hfil
  \subfloat[Art work at $2048 \times 2048$]{
    \includegraphics[width=.45\linewidth]{1k_gbp9047_70original.png} 
  }
  \subfloat[Activation]{
    \includegraphics[width=.45\linewidth]{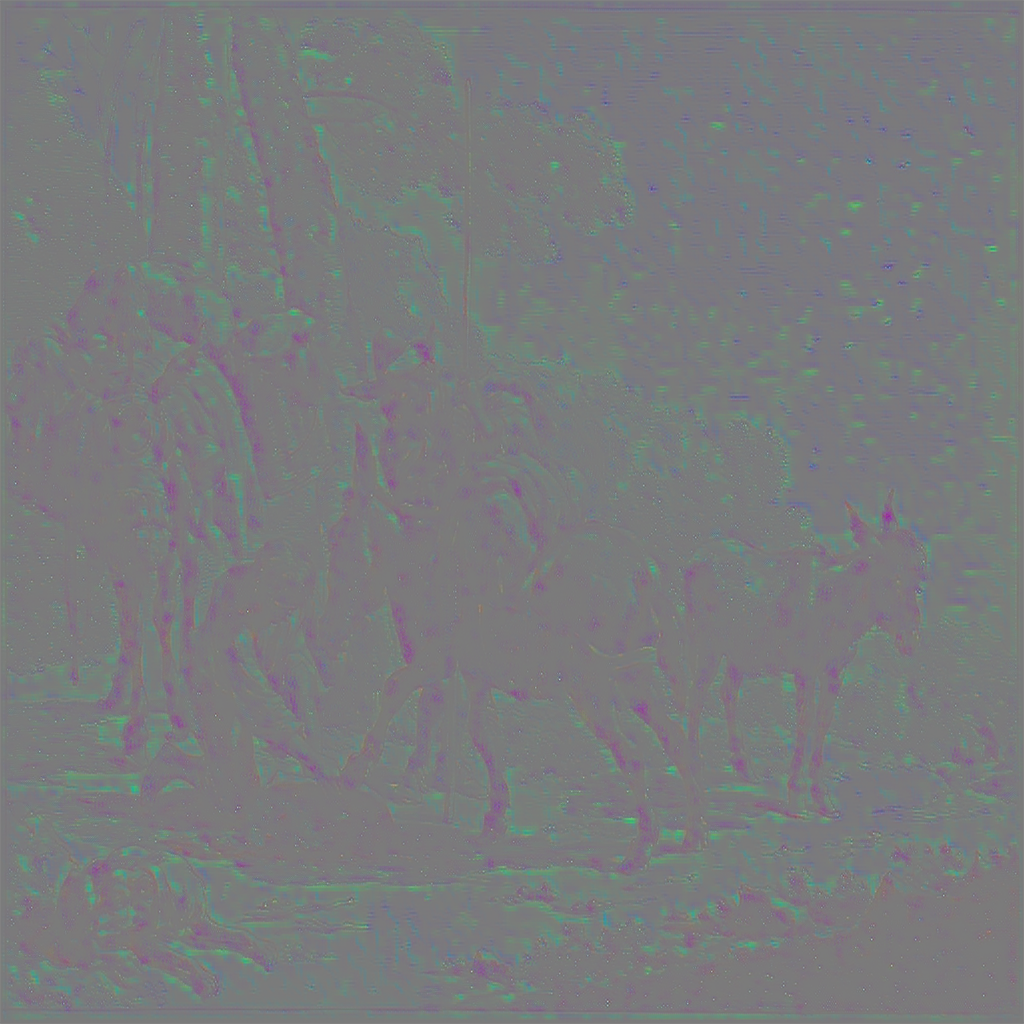}
  }
  \caption{Visualisations of the activations (right column) for the artwork \textit{`Hoefsmid bij een ezel'} by Jan de Visscher in the left column at four scales. The activation shows the importance of the highlighted regions for correctly identifying the artist. Best viewed in colour.}\label{fig:guidedbackp}
\end{figure}

To illustrate the effect of resolution on the automatic detection of artist-specific features, Guided Backpropagation \cite{springenberg2015} was used to create visualisations of the artwork \textit{`Hoefsmid bij een ezel'} by Jan de Visscher at at four scales. Figure~\ref{fig:guidedbackp} shows the results of applying Guided Backpropagation to the art work. The visualisations show the areas in the input image that the network considers characteristic of Jan de Visscher for that scale. A clear shift to finer details is observed when moving to higher resolutions.

\section{Discussion}
\label{sec:discussion}
The resolution at which artworks are investigated are usually decided based on the availability of the data and the computational cost of the analysis method. In this work we demonstrated, using a method that was designed to operate on images of $256 \times 256$ pixels, that the optimal scale at which artworks should be analysed for an artist attribution task differs per artist. By comparing the performance of a Convolutional Neural Network trained for artist attribution on images of artworks at $4$ different scales we find that on average it is beneficial to analyse images of artworks at a more detailed scale, but that for certain artists this has a negative impact on the classification performance.

On the dataset that was used to collect these results we found that found that the use of high-resolution images ($512 \times 512$ to $2048 \times 2048$ pixels) is beneficial to the artist attribution performance. When examining the performance for individual artists it became clear that there is a preferred scale for which the highest classification accuracy is achieved per artist. For the majority of the artists it was beneficial to analyse their artworks at a finer scale, but for almost $40\%$ of all artists it was more beneficial to analyse the images at a coarse scale. We suspect that this is due to (i) the fact that the images are not normalised to have a fixed pixel to mm ratio and (ii) compositional information being more salient on a coarse scale, thus making it easier to identify artists who have a very distinctive style in terms of composition or content matter. This also highlights another possible cause for these differences: the best performing scale for an artist is dependent on the other artists in the dataset. When the dataset consists of multiple artists who have a similar style on a fine scale,
but on a coarse scale they are very different, than it would be beneficial to analyse their works on a coarse scale. Because it is non-trivial to determine the relative optimal scale {\em a-priori} we pose that for future work it would be beneficial to focus on multi-scale approaches that exploit the discriminative information at multiple scales to attribute artworks accurately regardless of scale. Multi-scale image classification with CNNs has already been performed for object detection \cite{Sermanet2013} and traffic sign recognition \cite{Sermanet2011}. Additionally we suspect that more rigorous control of the image resolution in relation to the physical resolution can positively influence the training process.

Selecting patches at random during training can have the potential side-effect that certain input areas are never processed, making it possible that certain characteristics are never encountered during training. Because many of the visually distinctive characteristics in artworks are present throughout the artwork (e.g. the paper texture and tool marks) the chances of encountering an previously unseen characteristic becomes very slim after several epochs. For this reason we suspect that the effectiveness of this training procedure will be highly dependent on the input images, the task for which they are analysed, and the duration of training.

Although there is a lack of high resolution images of forgeries, on the basis of our results we expect that the reliability of forgery detection may be improved considerably by incorporating additional spatial detail. 

\section{Conclusion}
\label{sec:conclusion}
There is a vast amount visual information to be gleaned from multi-resolution images of artworks: clues about authenticity, indications of the used materials, traces of the hand of the master, and more. By analysing the large amount of available details in the images we were able to improve on the current state-of-the-art for computational artist attribution. Using neural networks and general purpose graphical processing hardware it has become possible to perform detailed computational analysis of artworks, vastly expanding the possibilities for computational art forensics.

\section*{Acknowledgment}
The research reported in this paper is performed as part of the REVIGO project, supported by the Netherlands Organisation for scientific research 
(NWO; grant 323.54.004) in the context of the Science4Arts research program. 

\bibliographystyle{IEEEtran}
\bibliography{library.bib}

\end{document}